\PassOptionsToPackage{dvipsnames,svgnames}{xcolor}
\documentclass[runningheads]{Latex/Classes/llncs}
\usepackage[
comment,
watermark,
highlightUnfinishContent,
Mathematics,
Logics,
BPMN,
]{Latex/Macros/Paper/paper}
\usepackage{Latex/Macros/macros_paper}

\usepackage{csquotes}

\usepackage[nonumberlist,xindy]{glossaries}
\providecommand*{\asnormaltext}[2][\@empty]{%
  \ifthenelse{\equal{#1}{\@empty}}%
    {\xspace{\normalfont{#2}}}%
    {\xspace{\normalfont{#1{#2}}}}%
  }
\providecommand*{\asnormaltextp}[2][\@empty]{\asnormaltext[#1]{(#2)}}

\newglossaryentry{a-compliant} {
 name={A-compliant},
 description={A-Compliant},
}
\newglossaryentry{activity} {
 name={activity},
 plural={activities},
 description={Activity},
}
\newglossaryentry{agreement} {
 name={agreement},
 description={agreement},
}
\newglossaryentry{annotated wf-net} {
 name={annotated \gls{wf-net}},
 plural={annotated \glspl{wf-net}},
 description={Annotated WF-net},
 descriptionplural={Annotated WF-nets},
}
\newglossaryentry{annotation} {
 name={annotation},
 description={Annotations},
}
\newglossaryentry{annotation function} {
 name={annotation function},
 description={Annotation Function},
}
\newglossaryentry{attribute} {
 name={attribute},
 description={Attributes},
}
\newglossaryentry{artefact} {
 name={artefact},
 description={Artefact},
}
\newglossaryentry{bm} {
 name={business model},
 description={Business Model},
}
\newglossaryentry{bp} {
 name={business process},
 text={BP},
 description={Business Process},
 descriptionplural={Business Processes},
 first={business process \asnormaltextp{BP}},
 firstplural={business processes \asnormaltextp{BPs}},
}
\newglossaryentry{bpc} {
 name={BPC},
 description={Business Process Compliance},
 first={business process compliance \asnormaltextp{BPC}},
 firstplural={business processes compliances \asnormaltextp{BPCs}},
}
\newglossaryentry{bpmn} {
 name={{BPMN}},
 plural={{BPMN}},
 description={BPMN},
}
\newglossaryentry{bpmnq} {
 name={\mbox{BPMN-Q}},
 plural={\mbox{BPMN-Q}},
 description={BPMN-Q},
}
\newglossaryentry{c-semiring} {
 name={c-semiring},
 description={constraint semiring},
 descriptionplural={constraint semirings},
 first={\glsentrydesc{c-semiring} \asnormaltextp{c-semiring}},
 firstplural={\glsentrydescplural{c-semiring} \asnormaltextp{c-semirings}},
}
\newglossaryentry{cmf} {
 name={CMF},
 description={compliances management framework},
 descriptionplural={compliances management frameworks},
 first={compliances management framework \asnormaltextp{CMF}},
 firstplural={compliances management frameworks \asnormaltextp{CMFs}},
}
\newglossaryentry{compas} {
 name={{COMPAS}},
 plural={{COMPAS}},
 description={{COMPAS}},
}
\newglossaryentry{compliant} {
 name={compliant},
 description={Compliant},
}
\newglossaryentry{compliant trace} {
 name={compliant trace},
 description={Compliant Trace},
}
\newglossaryentry{compliant process} {
 name={compliant process},
 plural={compliant processes},
 description={Compliant Process},
 descriptionplural={Compliant Processes},
}
\newglossaryentry{control-flow} {
 name={control-flow},
 description={Control-Flows},
}
\newglossaryentry{d-compliant} {
 name={D-compliant},
 description={D-Compliant},
}
\newglossaryentry{d-compliance} {
 name={D-compliance},
 description={D-Compliance},
}
\newglossaryentry{data} {
 name={data},
 plural={data},
 description={data},
}
\newglossaryentry{deliverable} {
 name={deliverable},
 description={Deliverables},
}
\newglossaryentry{declare} {
 name={Declare},
 plural={Declare},
 description={Declare},
}
\newglossaryentry{epc} {
 name={EPC},
 description={EPC},
}
\newglossaryentry{fcl} {
 name={FCL},
 plural={FCL},
 description={Formal Contract Language},
 descriptionplural={Formal Contract Language},
 first={\glsentrydesc{fcl} \asnormaltextp{FCL}},
 firstplural={\glsentrydescplural{fcl} \asnormaltextp{FCL}},
}
\newglossaryentry{fmi} {
 name={functional mock-up interface},
 description={Functional Mock-up Interface},
}
\newglossaryentry{full compliance} {
 name={full compliance},
 description={full Compliance},
}
\newglossaryentry{fully compliant} {
 name={fully compliant},
 description={fully Compliant},
}
\newglossaryentry{invisible task} {
 name={invisible task},
 description={invisible task},
}
\newglossaryentry{gateway} {
 name={gateway},
 description={Gateway},
}
\newglossaryentry{kpi} {
 name={KPI},
 description={key performance indicator},
 descriptionplural={key performance indicators},
 first={\glsentrydesc{kpi} \asnormaltextp{KPI}},
 firstplural={\glsentrydescplural{kpi} \asnormaltextp{KPIs}},
}
\newglossaryentry{input} {
 name={input},
 description={Input},
}
\newglossaryentry{input attribute} {
 name={input attribute},
 description={input attribute},
}
\newglossaryentry{labelled wf-net} {
 name={labelled WF-net},
 description={Labelled Workflow Net},
}
\newglossaryentry{literal} {
 name={literal},
 description={Literals},
}
\newglossaryentry{literal set} {
 name={literal set},
 plural={literal set},
 description={Literal Sets},
}
\newglossaryentry{m-compliant} {
 name={M-compliant},
 description={M-Compliant},
}
\newglossaryentry{moderately compliant} {
 name={moderately compliant},
 description={Moderately Compliant},
}
\newglossaryentry{occurrence sequence} {
 name={occurrence sequence},
 description={Occurrence Sequences},
}
\newglossaryentry{output} {
 name={output},
 description={Outputs},
}
\newglossaryentry{output attribute} {
 name={output attribute},
 description={output attributes},
}
\newglossaryentry{p-compliant} {
 name={P-compliant},
 description={P-Compliant},
}
\newglossaryentry{partial} {
 name={partial},
 description={partial},
}
\newglossaryentry{partial compliance} {
 name={partial compliance},
 description={Partial Compliance},
}
\newglossaryentry{partially compliant} {
 name={partially compliant},
 description={partially compliant},
}
\newglossaryentry{not-compliant} {
 name={not-compliant},
 description={not-Compliant},
}
\newglossaryentry{non-compliant} {
 name={non-compliant},
 description={Non-Compliant},
}
\newglossaryentry{process model} {
 name={process model},
 description={Process Models},
}
\newglossaryentry{regulatory compliance} {
 name={regulatory compliance},
 description={Regulatory Compliances},
}
\newglossaryentry{resource} {
 name={resource},
 description={Resources},
}
\newglossaryentry{seaflows} {
 name={\mbox{SeaFlows}},
 plural={\mbox{SeaFlows}},
 description={SeaFlows},
}
\newglossaryentry{strongly compliant} {
 name={strongly compliant},
 description={strongly Compliant},
}
\newglossaryentry{sub-compliant} {
 name={sub-compliant},
 description={sub-Compliant},
}
\newglossaryentry{t-compliance} {
 name={T-compliance},
 description={T-Compliance},
}
\newglossaryentry{t-compliant} {
 name={T-compliant},
 description={T-Compliant},
}
\newglossaryentry{task} {
 name={task},
 description={Tasks},
}
\newglossaryentry{task annotation} {
 name={task annotation},
 description={Task Annotations},
}
\newglossaryentry{task compliance measure} {
 name={task compliance measure},
 description={Task Compliance Measure},
}
\newglossaryentry{time} {
 name={time},
 description={times},
}
\newglossaryentry{trace} {
 name={trace},
 description={Traces},
}
\newglossaryentry{tuple} {
 name={tuple},
 description={tuples},
}
\newglossaryentry{ui} {
 name={user interface},
 description={user interface},
}
\newglossaryentry{visible task} {
 name={visible task},
 description={visible task},
}
\newglossaryentry{weakly compliant} {
 name={weakly compliant},
 description={Weakly Compliant},
}
\newglossaryentry{wf-net} {
 name={WF-net},
 description={Workflow Net},
 descriptionplural={Workflow Nets},
 first={\glsentrydesc{wf-net} \asnormaltextp{WF-net}},
 firstplural={\glsentrydescplural{wf-net} \asnormaltextp{WF-nets}},
}
\newglossaryentry{transition} {
 name={transition},
 description={transition},
}

\makeglossaries

\def\papertitle{Towards a Formal Framework for Partial Compliance of Business Processes}

\usepackage{cleveref}

\begin{document}
\title{\papertitle}
\author{Ho-Pun Lam\inst{1} %\orcidID{0000-0002-1137-8549}
\and
Mustafa Hashmi\inst{1,2,3}   %\orcidID{0000-0002-6376-082X}
\and
Akhil Kumar\inst{4} % \orcidID{2222--3333-4444-5555}
}
\authorrunning{Lam, Hashmi and Kumar}
%\authorrunning{H.-P. Lam et al.}
\institute{Data61, CSIRO, Australia \and
Federation University, Brisbane QLD 4000, Australia \and
La Trobe Law School, La Trobe University, Melbourne, Australia \\
\email{\{brian.lam,mustafa.hashmi\}@data61.csiro.au}\\
\and
% Department of Supply Chain and Information Systems,
Smeal College of Business, Penn State University, University Park, PA 16802, USA\\
\email{akhilkumar@psu.edu}}
\maketitle              % typeset the header of the contribution
\begin{abstract}
%!TEX root = ../manuscript.tex

%Due to circumstances
%that may be %possibly 
%beyond our control,
%or lack of information 
%%exists 
%to assure safety
%and effectiveness of the business operations,
%practitioners,
%or compliance managers,
%often assess whether a business process is compliant with the intended outcomes.
%Though this can clarify whether the intended result was received by target audiences,
%practitioners often like to go a step further
%and identify the issues from the samples to minimize the disruption
%that could possibly occur.
% Besides,
% they also interested in knowing more than a binary information on the state-of-affairs
% of the compliance for a better understanding of the potential risks due to violations
% and other detrimental effects
% so that they can identify the source of the problems
% and allocate enough resources handle the non-compliance issues
% before they appear.
Binary ``YES-NO'' notions of process compliance are not very helpful to managers for assessing the operational performance of their company because 
a large number of cases fall in the grey area of partial compliance.  
Hence, it is necessary to have ways to quantify partial compliance in terms of metrics and be able to classify 
actual cases by assigning a numeric value of compliance to them.  
In this paper,
we formulate an evaluation framework to quantify the level of compliance of business processes
across different levels of abstraction (such as task, trace and process level) and across multiple dimensions 
of each task (such as temporal, monetary, role-, data-, and quality-related) to
provide managers more useful information about their operations and to help them improve their decision making processes.
Our approach can also add social value by making social services provided by local,
state
and federal governments more flexible and improving the lives of citizens.

\end{abstract}

%!TEX root = manuscript.tex

\glsresetall

%!TEX root = ../manuscript.tex

\section{Introduction}
\label{section:introduction}

% \Gls{regulatory compliance} aims to ensure
% that enterprises' business operations are in alignment with the legislations
% or regulation promulgated by legitimate authority.
% The requirements for being compliant has increase over the last two decades due to the big corporate scandals,
% such as Enron (\SI{74}[\$]{\billion}),
% WorldCom (\SI{180}[\$]{\billion}),
% American International Group (\SI{3.9}[\$]{\billion}),
% Bernnie Madoff (\SI{65}[\$]{\billion}),
% Lehmann Brothers (\SI{50}[\$]{\billion}),
% in America,
% and Soci\'{e}t\'{e} G\'{e}n\'{e}rale (\SI{4.9}[\EURO]{\billion}),
% UBS (\SI{2.3}[\EURO]{\billion}) in Europe.
% Such big loses,
% in some cases,
% leading to the closures of large companies,
% resulted in the need to design
% and implement new regulations to regulate how business should operate.
% Subsequently,
% several new regulations,
% such as BASEL~\cite{BAS13},
% HIPAA~\cite{HIPAA96},
% and Sarbanes-Oxlay~\cite{US02},
% have been emerged,
% imposing severe penalties,
% in both financial and/or criminal,
% for non-compliance,
% which bestow more pressure on enterprises to streamline their operations to remain compliant.

When designing \glspl{bp},
practitioners always assume
that the \gls{bm} will be executed as planned.
However,
this is impractical in many situations.
For example,
cost fluctuations,
% the cost fluctuation may have significant impact on the objective of the process,
equipment and resource availability,
time constraints, and
human errors can cause disruptions.
In response to this,
it is crucial for the practitioners to have a complete picture
of the status of their running
business processes %\glspl{bp}
\textemdash\xspace for taking strategic decisions on identifying,
forecasting,
obtaining
and allocating required resources,
and to be notified
if any non-compliance issues are identified during execution.

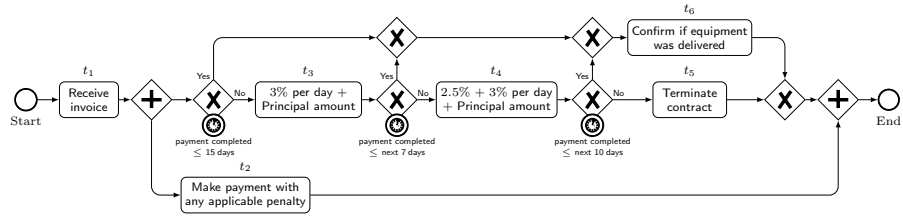
\begin{figure*}[t]
% \vspace*{-.6em}
% \ignore{\resizebox{\linewidth}{!}{\input{images/frag2}}}
\resizebox{\linewidth}{!}{%!TEX root = ../paper.tex

{
\begin{tikzpicture}[
ProcessNode/.style={processNode,font=\sffamily\scriptsize,rounded corners=3pt,text width=3em,minimum height=2.2em},
GatewayNode/.style={gatewayNode,minimum width=2.5em,minimum height=2.5em},
EventNode/.style={eventNode},
EventSymbol/.style={eventNode,yshift=-.3em,fill=white,minimum width=1.3em,minimum height=1.3em},
EventSymbolLabel/.style={anchor=north,draw=none,font=\sffamily\tiny,text centered,text width=4.7em,inner sep=2pt,yshift=1pt},
placeHolderNode/.style={draw=none,minimum width=#1,opacity=0},
taskId/.style={draw=none,font=\scriptsize,text centered,inner sep=3pt,minimum height=0em},
lbl/.style={Label,font=\sffamily\tiny},
conditionLabel/.style={lbl,yshift=-1.9em,text width=30,anchor=north,below},
decisionLabel/.style={lbl,text centered},
l/.style={->,rounded corners=5pt,shorten >=0pt,shorten <=0pt},
]

\matrix (base) [draw=none,
  column sep=7pt,row sep=.2em,
  inner sep=0pt,
] {
 & & & [.4em] & [.4em] & \node (gate1) [GatewayNode,ExclusiveGateway] {};
  & [.25em] & \node (gate2) [GatewayNode,ExclusiveGateway] {};
  & \node (pConfirm) [ProcessNode,text width=7.2em,label={[taskId]:\task[6]}] {Confirm if equipment was delivered} ;
  \\
\node (start) [EventNode,StartEvent,label={[taskId]below:Start}] {}; 
  & \node (p1) [ProcessNode,text width=3.2em,label={[taskId]:\task[1]}] {Receive invoice};
  & \node (gate7) [GatewayNode,ParallelGateway] {}; 
  & \node (gate3) [GatewayNode,ExclusiveGateway] {}; \node [placeHolderNode={2.8em}] {}; 
  & \node (p3) [ProcessNode,text width=6.1em,label={[taskId]:\task[3]}] {\SI{3}{\!\percent} per day + Principal amount};
  & \node (gate4) [GatewayNode,ExclusiveGateway] {}; \node [placeHolderNode={2.8em}] {}; 
  & \node (p4) [ProcessNode,text width=7em,label={[taskId]:\task[4]}] {\SI{2.5}{\!\percent} + \SI{3}{\!\percent} per day\par + Principal amount};
  & \node (gate5) [GatewayNode,ExclusiveGateway] {}; \node [placeHolderNode={2.8em}] {}; 
  & \node (p5) [ProcessNode,text width=4.1em,label={[taskId]:\task[5]}] {Terminate contract}; 
  & \node (gate3e) [GatewayNode,ExclusiveGateway] {};
  & \node (gate8) [GatewayNode,ParallelGateway]{};
  & \node (end) [EventNode,EndEvent,label={[taskId]below:End}] {};
  \\
};

\foreach \n/\l [count=\c] in {gate3/\SI{15}{days},gate4/next~\SI{7}{days},gate5/next~\SI{10}{days}} {
  \node (eventSymbol-\c) [EventSymbol,TimerStartEvent] at (\n.south) {} ; 
  \sisetup{detect-family}
  \node (eventSymbol-\c-label) [EventSymbolLabel] at (eventSymbol-\c.south) {payment completed\\ $\mathsf{\tiny\leq\l}$} ; 
}

\node (p6) [ProcessNode,anchor=north west,text width=7.5em,xshift=-.75em,yshift=-1em,label={[taskId]:\task[2]}]
  at (eventSymbol-1-label.south west-|gate3.west) {Make payment with any applicable penalty}; 

\foreach \n [remember=\n as \lastn (initially start)] 
    in {p1,gate7,gate3,p3,gate4,p4,gate5,p5,gate3e,gate8,end} {
  \draw[l] (\lastn) -- (\n) ;
}

\draw [l] (gate1) -- (gate2);

\draw [l] (gate3) |- (gate1) ;
\draw [l] (gate2) -- (pConfirm) ;
\draw [l] (pConfirm) -| (gate3e) ;

\foreach \na/\nb in {gate4/gate1,gate5/gate2} {
  \draw[l] (\na) -- (\nb) ;
}
\draw [l] (gate7) |- (p6) ;
\draw [l] (p6) -| (gate8) ;

\foreach \n in {gate3,gate4,gate5} {
  \node [decisionLabel,anchor=south east,yshift=-2pt] at (\n.north) {Yes} ;
}
\foreach \n in {gate3,gate4,gate5} {
  \node [decisionLabel,anchor=south west,xshift=-2pt] at (\n.east) {No} ;
}

\end{tikzpicture}
}}
\caption{Fragment of payment-making process \adoptedfromp{\cite{Hashmi.2015}}}
\label{figure:fragmentOfPaymentProcessModel}
% \vspace*{-1em}
\end{figure*}

Let us illustrate this idea
by examining the payment \gls{process model} as shown in \Cref{figure:fragmentOfPaymentProcessModel},
which consists of a sequence of tasks to be performed.
% which depicts a fragment of payment-making process.
Accordingly,
a customer is required to make the payment within \SI{15}{\days} upon receiving the invoice;
if not,
the invoice must be paid with \SI{3}{\!\percent} per day interest in addition to
principal amount within the next \SI{7}{\days}.
For any subsequent days hereafter within the next \SI{10}{\days},
an additional \SI{2.5}{\!\percent} interest will be added to the total payment as penalty, which will be
calculated based on the principal amount.
The contract will be terminated automatically upon $3$ consecutive defaults.

Now, consider two compliant executions performed by two business customers of the company,
customers A and B.
Customer A strictly follows the \emph{normal} sequence
and makes the payment within \SI{15}{\days} after receiving the invoice.
Customer B instead delayed the payment
and paid the bill (together with interest and penalty) \SI{3}{\weeks} later.
If we ascribe value to this process depending on the billing company's revenue,
both executions positively contribute to it,
as both customers did make their payments after receiving the invoices.
However,
the deferred payment of company B may affect the cash flow of the service provider company. Moreover, both these scenarios represent examples of partial compliance because there was a violation on the temporal dimension.  Other violations may occur along other compliance dimensions such as: \emph{money}, when monetary payments are not made according to agreements; \emph{roles}, when individuals who perform certain tasks like approvals, etc., are not in the normal or authorized, or delegated, role; \emph{data}, when the complete data required to perform a task is not available; and \emph{quality}, when the quality of the work performed by a task is sub-standard.  For each dimension, there are prescribed ranges of values or performance indicators in which a task is considered to be compliant on that dimension.  If the indicator values  within a narrow range are outside this normal range, then the task is said to be in partial compliance on that dimension. Finally, if the indicator does not fall into either of these two ranges then it is said to be non-compliant. A dimension can also be related to an attribute value. Thus, $payInDays$ attribute represents the number of days within which payment is made after the invoice is sent to the customer. This attribute corresponds to the temporal dimension and can be used interchangeably with it.

\fussy
Existing systems and compliance management frameworks
(such as \gls{declare}~\cite{Maggi.2011},
\sloppy
\gls{seaflows}~\cite{Ly.2012},
\gls{compas}~\cite{Schumm.2010a},
etc.)
only provide an \emph{all{-}or{-}nothing} type of binary answer, i.e.,
\emph{YES} if the \gls{bp} is \emph{fully} compliant;
and \emph{NO} if any non-compliant behavior has been detected at some point during execution,
which is not \emph{informative}
and raises a simple yet significant question of
whether the \emph{whole} process is not compliant
or \emph{only} a part of it,
and whether  corrective actions should be performed from the point
where the non-compliant behavior was detected,
or from an earlier point.

Recently,
some efforts coining the notion of \emph{\gls{partial compliance}} have been reported.
% \footnote{Note that the definition
%of \gls{partial compliance} in this paper is different from that in legal domain.
%Legally speaking partial compliance is not compliant. Let’s take legal perspective into account by referring clause \textsf{\S73.4 Partial compliance not deemed compliance:} The fact that a notification has been filed shall not necessarily be deemed full compliance with \textsf{\S18 U.S.C.951} or these regulations on the part of the agent; nor shall it indicate that the Attorney General has in any way passed on the merits of such notification or the legality of the agent’s activities; nor shall it preclude prosecution, as provided for in \textsf{\S18 U.S.C.951}, for failure to file a notification when due, or for a false statement of a material fact therein, or for an omission of a material fact required to be stated therein (\url{https://www.law.cornell.edu/cfr/text/28/73.4}).
%
%In this paper, we differentiate partial compliance in the context of level of compliance in a business process
%when some of the tasks are executed under some sub-ideal situation.
%More details will be provided in \Cref{section:framework}.}
For example, the approach in
\cite{Lu.2008a} returns the status of a \gls{bp} as \emph{ideal},
\emph{sub-ideal},
\emph{non-compliant}
and \emph{irrelevant}.
Based on the notion of decision lattices~\cite{Huchard.2007},
\etal{Morrison}\cite{Morrison.2009} categorizes the compliance status as \emph{Good},
\emph{Ok},
and \emph{Bad}.
However,
the issues remain similar as:  \emph{to what extent the process is compliant
and how much (or what kind of) additional resources are required to resolve
any detected non-compliance issues?}

To answer this question,
in this paper,
we present a formal framework for evaluating the levels of compliance of a \gls{bp} at different levels of abstraction during execution
and auditing phases,
% and auditing phases~\cite{Hashmi.2018},
aiming to provide more clear
and useful information to users concerned in facilitating their decision making process
when any non-compliance issues arise.

The rest of the paper is organized as follows.
Next, in \Cref{section:background}, we 
provide necessary background information and 
terminologies following which we introduce 
our proposed framework in \Cref{section:framework}. 
Examples illustrating how the proposed 
framework works in practice are 
presented in \Cref{section:exampleOfMeasureCalculation}. 
Related work is discussed in~\Cref{section:relatedWork}
before the paper is concluded 
with final remarks 
and directions for future work in \Cref{section:conclusions}.

%present the necessary background information
% and terminology used for understanding the proposed framework.
% Then,
% our proposed framework will be presented in \Cref{section:framework},
% followed by examples illustrating how the framework works in practice.
% \Cref{section:relatedWork} tackle related work
% and \Cref{section:conclusions} concludes the paper
% and provide directions for future works.

% \Cref{section:relatedWork} presents the related work;
% while \Cref{section:background} introduces some of the necessary background
% and terminology for the understanding our proposed frameworks.

% \Cref{section:introduction}: Introduction

% \Cref{section:background}: Background and Problem Statement

% \Cref{section:framework}: \glsdesc{partial compliance} Framework

% \Cref{section:relatedWork}: Related Work

% \Cref{section:conclusions}: Conclusions

%!TEX root = ../manuscript.tex

\section{Background and Problem Statement}
\label{section:background}

In this section,
we first introduce the necessary background
and terminologies for the understanding of our proposed framework,
and subsequently derive the problem statement.

\subsection*{Structure of a \Glsdesc{bp}}

\begin{figure}[t]\centering
%!TEX root = ../manuscript.tex

{
\begin{tikzpicture}[
ProcessNode/.style={processNode,inner sep=2pt},
unusedDeliverables/.style={decorate,decoration={coil,aspect=0,pre=lineto,pre length=1pt,post length=3pt}},
resourcesLine/.style={auto,decorate,decoration={shape backgrounds,shape=dart,shape sep=5pt},shape border rotate=180},
linelabel/.style={LineLabel},
]

% \node (rules) [font=\scriptsize,draw,inner sep=4pt,rounded corners=3pt] {Rules} ;
% \node (task) [anchor=south,inner sep=3pt] at ($(rules.north)+(0pt,2pt)$) {$Task_i$} ;
\node (task) [anchor=south,inner sep=3pt,minimum height=1.8em] {$Task_i$} ;

\node (task-1) [ProcessNode,fit=(task),minimum width=5.4em,inner sep=3pt] {};

\coordinate (task-0) at ($(task-1.west)+(-6em,0)$) ;
\coordinate (task-2) at ($(task-1.east)+(6em,0)$) ;

\foreach \n [count=\c, remember=\c as \lastc (initially 0)] in {i-1} {
  \draw [->,dashed,yshift=-10pt] ($(task-\lastc.east)+(0,5pt)$) to [out=20,in=160] ($(task-\c.west)+(0,5pt)$)
  	;
  \draw [->] (task-\lastc) -- (task-\c)
    node (effect-\c) [linelabel,swap,midway,yshift=3.0em] {$\attribute[\n][1..n_{\n}]$} 
    ;
}

\foreach \n/\lastc/\c in {i/1/2} {
  \draw [->,dashed,yshift=-10pt] ($(task-\lastc.east)+(0,-5pt)$) to [out=340,in=200] ($(task-\c.west)+(0,-5pt)$)
    ;
  \draw [->] (task-\lastc) -- (task-\c)
    node (effect-\c) [linelabel,swap,midway,yshift=-1.6em] {$\attribute[\n][1..n_{\n}]$} 
    ;
}

\draw [->,unusedDeliverables] (task-1) 
  -- ++ (4.95em,3em) 
    node [linelabel,inner sep=1pt,auto,anchor=south,xshift=-0em] {$\deliverable[i][1..d_{i}]$}
    ;

% \draw [<-,resourcesLine] (task-1.south) + (0,-5pt) 
%   -- ++ (0,-1.9em) 
%     node [linelabel,anchor=north,inner sep=2pt,yshift=5pt] {$\resource[i][1..r_{i}]$}
%     ;

\end{tikzpicture}
% \medskip
}
\caption{\Glspl{annotation} of a \gls{task}}
\label{figure:taskAnnotations}
\end{figure}
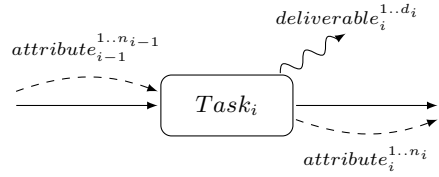

A \gls{bp} is represented as a
temporally
and logically ordered, directed graph in which the nodes
represent tasks of the process that are executed to achieve a specific goal.
It describes what needs to be done
and when (\emph{\glspl{control-flow}} and \emph{\gls{time}}),
who is involved (\emph{\glspl{resource}}),
and what it is working on (\emph{\gls{data}})~\citep{Governatori.2013}.
Essentially,
a \gls{bp} is composed of various elements
which provide building blocks for aggregating loosely-coupled (atomic) \emph{\glspl{task}}
% (or \emph{\glspl{task}})
as a sequence in a process aligned with the business goals.

% \begin{wrapfigure}{r}{.475\linewidth}\centering
% \vspace{-1em}
% \resizebox{\linewidth}{!}{\input{images/task2}}
% \caption{\Glspl{annotation} of a \gls{task}}
% \label{figure:taskAnnotations}
% \vspace{-2em}
% \end{wrapfigure}

% \begin{figure}[t]\centering
% \input{images/task2}
% \caption{\Glspl{annotation} of a \Gls{task}}
% \label{figure:taskAnnotations}
% \end{figure}

Each \gls{task} is  an atomic unit of work
with its own set of \emph{\glspl{input attribute}},
which can be (partially) aggregated from \emph{preceeding} \gls{task}(s)
or acquired from other {sources},
describing the \emph{prerequisites}
or \emph{requirements}
that the \gls{task} has to comply with for its (full) execution,
and \emph{\glspl{output attribute}}
that it has to produce upon execution
and for propagation to \emph{succeeding} \glspl{task} (as input),
as shown in \Cref{figure:taskAnnotations}.  Note that,
in \Cref{figure:taskAnnotations},
the term \emph{\gls{deliverable}} is used to describe
an output document or \emph{\gls{artefact}} that is produced by the task
after execution and is not propagated to the next task.
Technically,
the values of \glspl{attribute} (both input and output) can have multiple
dimensions,
which may include information about time (or \emph{temporal}),
monetary, data, role, quality of service,
or any combination of these.
As an example,
the value of payment
and the payment due date of \task[2] in \Cref{figure:fragmentOfPaymentProcessModel}
are from temporal
and monetary dimensions,
respectively.

% Notice here
% that inputs can be \emph{\glspl{attribute}}
% (or \emph{information}, including normative effects) aggregated from preceeding \glspl{task}),
% or data or resources that are external from the process;
% while outputs are attributes that are generated after \gls{task} execution
% and will be propagated to succeeding \gls{task},
% or non-propagated \emph{\glspl{artefact}} (or \emph{\glspl{deliverable}},
% such as reports) produced by the \gls{task}.

A sequence representing the execution order of \glspl{task} of a \gls{bp}
in a given case is called a \gls{trace} (a.k.a. \gls{occurrence sequence}).
Typically,
a \gls{bp} can be executed in a number of ways.
For instance,
below is the set of \glspl{trace}
that can be generated from the \gls{bm},
from \emph{start} to the \emph{end},
as shown in \Cref{figure:fragmentOfPaymentProcessModel}.
% \cq{not quite sure what that mean}{However,
% no matter in what order a \gls{bp} is executed,
% its execution delivers value(s) to all the entities involved in the process.}

\smallskip
% \textcolor{red}{
{\small\[
\begin{array}{l@{}l@{\qquad}l@{}l}
\traceset=\{
  & \trace[1]=\angleset{\transition[1],\transition[2],\transition[3],\transition[4],\transition[5]},
  & \trace[2]=\angleset{\transition[1],\transition[3],\transition[2],\transition[6]},
  \\
  & \trace[3]=\angleset{\transition[1],\transition[2],\transition[3],\transition[4],\transition[6]},
  & \trace[4]=\angleset{\transition[1],\transition[3],\transition[4],\transition[2],\transition[5]},
  \\
  & \trace[5]=\angleset{\transition[1],\transition[2],\transition[3],\transition[6]},
  & \trace[6]=\angleset{\transition[1],\transition[3],\transition[4],\transition[2],\transition[6]},
  \\
  & \trace[7]=\angleset{\transition[1],\transition[2],\transition[6]},
  & \trace[8]=\angleset{\transition[1],\transition[3],\transition[4],\transition[5],\transition[2]},
  \\
  & \trace[9]=\angleset{\transition[1],\transition[3],\transition[2],\transition[4],\transition[5]},
  & \trace[10]=\angleset{\transition[1],\transition[3],\transition[4],\transition[6],\transition[2]},
  \\
  & \trace[11]=\angleset{\transition[1],\transition[3],\transition[2],\transition[4],\transition[6]},
  & \trace[12]=\angleset{\transition[1],\transition[3],\transition[6],\transition[2]},
  \\
  & \trace[13]=\angleset{\transition[1],\transition[6],\transition[2]}
  \} \\
\end{array}
\]}
% }
\smallskip

\begin{figure*}[t]\centering
%!TEX root = ../paper.tex

{
\def\offset{3pt}
\begin{tikzpicture}[
font=\sffamily\footnotesize,
every node/.style={inner sep=2pt},
% triangleNode/.style={regular polygon,regular polygon sides=3,minimum width=#1},
triangleNode/.style n args={2}{isosceles triangle,minimum width=#1,minimum height=#2,isosceles triangle stretches},
termNode/.style={solid,inner sep=2pt,font=\sffamily\small\bfseries},
actNode/.style={font=\sffamily\footnotesize,inner sep=0pt,inner xsep=3pt},
lbl/.style={Label,inner sep=2pt,font=\sffamily\scriptsize},
alternativePath/.style={dashed},
]

\node (triangle) [Wrapper,triangleNode={7.5em}{11em},rotate=180] {} ;

\node (violation) [termNode,anchor=east] at (triangle.east) {Violation} ;
\node (compensation) [termNode,anchor=south] at (triangle.right corner) {Compensation} ;
\node (penalty) [termNode,anchor=north] at (triangle.left corner) {\begin{varwidth}{\linewidth}\centering Penalty/\\ Punishment\end{varwidth}} ;

\draw [->,alternativePath] ($(violation.east)+(0,-\offset)$)
  -- node [lbl,pos=.45,swap,yshift=1pt] {\begin{varwidth}{\linewidth}\centering induces/\\ demands\end{varwidth}}
    ($(penalty.north)+(-\offset,.5\offset)$) ;

\draw [->] ($(violation.east)+(0,\offset)$)
  -- node [lbl] {seeks}
    ($(compensation.south)+(-\offset,-.5\offset)$) ;

\draw [->,alternativePath] ($(compensation.south)+(\offset,0)$)
  -- node [lbl] {\begin{varwidth}{\linewidth}incur/\\ subject to\end{varwidth}}
    ($(penalty.north)+(\offset,0)$) ;

\node (terminateProcess) [actNode,inner ysep=1pt] at (violation|-penalty) {\begin{varwidth}{\linewidth}\centering Terminate/Suspend\\ Process\end{varwidth}} ;

\draw [->,dashed] (violation)
  -- node [lbl,swap] {leads to}
    (terminateProcess) ;

\draw [->,dashed] (penalty)
  -- node [lbl] {can lead to}
    (terminateProcess) ;

\node (continuesExceution) [actNode,xshift=1.9em,anchor=west]
  at (compensation.east) {\begin{varwidth}{\linewidth}\centering{Continues\\ Execution}\end{varwidth}} ;
\node (normalTermination) [actNode,xshift=1.8em,anchor=west]
  at (continuesExceution.east) {\begin{varwidth}{\linewidth}\centering{Normal\\ Termination}\end{varwidth}} ;

\draw [->,alternativePath] (compensation) -- (continuesExceution) ;
\draw [->,alternativePath] (continuesExceution) -- (normalTermination) ;

\draw [->,alternativePath] ($(penalty.east)+(-1pt,0pt)$)
  to [out=0,in=-93]
    node [lbl,swap] {resumes}
    (continuesExceution) ;

% \draw [->] (violation)
%   to [out=-50,in=180]
%     node [pos=.5,lbl,swap] {induce/demand} 
%     (penalty);

\end{tikzpicture}
}
\ifspringer\medskip\fi
% \resizebox{\linewidth}{!}{\input{images/violationCompensationRelation2}}
\caption{Violation-compensation relationships of a partially compliant \glsname{bp}}
\label{figure:violationCompensationPattern}
\end{figure*}
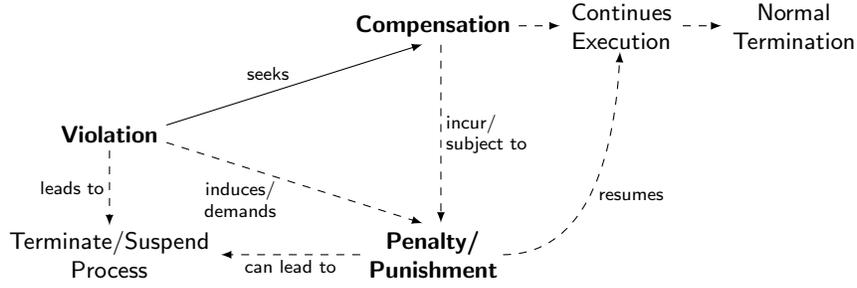

While it is always desirable
that a \gls{bp} behave strictly in accordance with the prescribed conditions,
this may not always be  the case in practice.
% It is possible
% that a process may deviate from its desired behavior
A \gls{bp} may deviate from its desired behavior
in unforeseen circumstances
and violate some (or all) of the conditions attached to it during execution.

% \begin{figure*}\centering
% \input{images/violationCompensationRelation2}
% % \resizebox{\linewidth}{!}{\input{images/violationCompensationRelation2}}
% \caption{Violation-compensation relationships of a partially compliant \glsname{bp}}
% \label{figure:violationCompensationPattern}
% \end{figure*}

\Cref{figure:violationCompensationPattern} illustrates
what can happen when a violation occurs in a \gls{bp}.
The divergent behavior % violating the process conditions
may cause a temporary suspension or (in some cases) termination of the process,
and may also induce penalties.
A \emph{penalty} is a punitive measure (e.g. monetary or in some other form)
enforced by company policy or a rule for the performance of
an action/act that is proscribed,
or for the failure to carry out some required acts.
%It entails the notion of punishment as pecuniary penalty. (e.g., financial, resources)
%or impose punishments (imprisonment).
However,
a violation of (mandatory) conditions does not necessarily imply automatic termination/suspension of a \gls{bp}
that would prevent any further execution.
%or impossibility of continue executing the \gls{bp}.
Certain violations can be compensated for~\citep{Hashmi.2016},
where compensation can be broadly understood as a remedial measure taken to offset the damage
or loss caused by the violation.
% ,as depicted in \Cref{figure:complianceDimensionsCompensationMechanisms}.
In general,
legal acts
and contracts provide clauses prescribing penalties
and remedial provisions
which are triggered
when the deviations from the contractual clauses occur.
These provisions may prescribe conditions
that are subject to some penalties
or punishments.
As mentioned in the previous section,
an execution with compensated violations (as in Figure 1) leads to a sub-ideal situation
~\citep{Sadiq.2007a}, and is deemed partially-compliant.
% and can continue executing normally once the compensatory measures are performed,
% but is executing under some \emph{sub-ideal} situation.
Nonetheless, the process can continue execution and complete normally
once the compensatory actions are performed.

Next,
we develop our framework in a formal manner.

\section{Partial Compliance Framework}
\label{section:framework}

% % Before answering this question,
% \Cref{figure:violationCompensationRelations} shows the consequence
% that can happen
% when a violation appear in a \gls{bp}.
% \toupdate{\begin{itemize}
% \item description of the image
% \item difference between compensation and penalty/punishment
% \item discussion with different domains, i.e., time, quality, data, etc.
% \item \ldots
% \end{itemize}}
In this section we develop a partial compliance framework. The framework is based on the following principles or axioms underlying partial compliance:

\begin{enumerate}[label={\bfseries Axiom \arabic*.},leftmargin=*]
\item Compliance should not be binary 0/1 but should cover a spectrum of scenarios between 0 and 1.

\item \Gls{partial compliance} should be recognized and treated fairly.

\item \Gls{partial compliance} can be rectified by compensation mechanisms such as imposition of penalties,
  or sanctions that increase monotonically with the extent of the violation.

\item The level of \gls{partial compliance} decreases monotonically as the magnitude of the violation increases.  
\end{enumerate}

Throughout this section,
we use the following notations:
$\taskidset$ is the set of unique \gls{task} identifiers of \glspl{task}
that appear in an instance of a \gls{trace} $\trace$;
and,
$\attributenameset$  denote the set of attribute names of task $t$.  
%and $\attributevalueset$
Each attribute is mapped to a value $v$ from a suitable numeric or categorical domain in a running instance.
Thus,
$(a,v)$ is a attribute-value pair or \gls{tuple} for an attribute in a task.

We introduce a partial compliance function $\evaluationprojectionfunctionsimple$
on task $\task$ to define partial compliance values for different attribute values under various compliance dimensions. 
Thus, $\evaluationprojectionfunction{\task}{a,v,d}$ denotes the degree of partial compliance of attribute $a$ of task $t$
where the value of attribute $a$ is equal to $v$,
on compliance dimension $d$. 
% on task $\task$ to define partial compliance values for different dimensions or attribute values.
This function maps attribute values of a task to a real-value in the \range{0}{1} range
that represents the degree of partial compliance, where 1 corresponds to full compliance.
Thus,
to formally describe the partial compliance for the running example in Figure 1, we can
write, $\evaluationprojectionfunction{\task[2]}{payInDays,\SI{10}{\days},Time}=1$.
This means that the partial compliance of task $T_2$ in dimension $Time$ is equal to 1
if the $payInDays$ attribute has a value of 10 days,
which also represents compliance of task $T_2$ on the temporal dimension.
%$\attributeprojectionfunction{\task}{a}=\attributemappingfunction{t}{a}$
%and $\evaluationprojectionfunction{\task}{a,v}=\evaluationfunction{t}{a}{v}$,
%where $a\in\attributenameset$ and $v\in\attributevalueset$.
%Hence,
%from \Cref{table:attributesMetric},
%$\attributeprojectionfunction{\task[2]}{payInDays}$ to retrieve the value of attribute $payInDays$ from the task $\task[t_2]$,
%and

%If, for a \gls{task} \task,
%$\attributemappingfunctionsimple$ does not contain any mapping to a particular attribute name,
%then the default value 1 is returned,
%i.e., $\evaluationprojectionfunction{\task}{\bot}{}=1$.

Given a \gls{task} $\task$,
we denote metric $\attributedimensiondefinition[\task]$ as the set of compliance dimensions
that relate to  $\task$  attributes
%\footnote{Note here that an attribute can be categorized into more than one dimension such as temporal, data, resource etc.}
% we denote the set of attributes dimensions as $\attributedimensiondefinition$
and denote its size as $\sizeof{\attributedimension}$.
Thus,
for the running example of Figure 1,
$\attributedimension_{\task[2]}=\set{\text{Monetary},$ $\text{Time},$ $\text{Percent}}$
and $\sizeof{\attributedimension}=3$.

% $\evaluationprojectionfunction{\task}{a}=\evaluationfunction{a}$

% and $quantizationvaluesimple$ on \task to extract specific attribute
% and quantification value from $\task$, respectively.
% That is,
% $\attributeprojectionfunction{\task}{a}=\attributemappingfunction{a}$
% and $quantizationvalue{\task}{a}=quantifyingfunction{a}$.
% Hence,
% from \Cref{table:attributesMetric},
% $\attributeprojectionfunction{\task[2]}{value}={100}$
% and \cq{to be updated}{$quantizationvalue{\task[2]}{value}=??$}

% for a task $\task=(\task[2],value={100},action={\text{make payment}},...)$
% $\projectfunction{\task}{value}=\text{100}$
% and $\projectfunction{\task}{action}=\text{make payment}$

% \begin{figure}\centering
% \input{images/violationCompensationRelation}
% % \resizebox{.8\linewidth}{!}{\input{images/violationCompensationRelation}}
% \caption{Violation-Compensation Relation}
% \label{figure:violationCompensationRelations}
% \end{figure}

Hence,
given a set of attribute names $\attributenameset$,
it is necessary to determine
which attributes relate to compliance
and  aggregate their individual compliance into a single metric of compliance.  Thus,
one can decide if a \gls{task} is fully-, partially- or non-compliant.
Accordingly,
we introduce the following definitions:

\begin{definition}[Aggregate attribute compliance  metric]\label{definition:dimensionIndex}
Given a \gls{task} $\task\in\taskidset$,
% $\attributenameset$ be the set of attribute names that it contains,
$\attributedimensiondefinition[\task]$ as the set of $n$ compliance dimension(s)  of  attributes of \task,
and an attribute aggregation operator $\attributeaggregationoperator$, then
we define a compliance metric for  attribute $a$ of task $t$ on dimension $i$ be:
% we define a compliance metric for  attribute $a$ of task $t$ on dimension $i$, where $a = v$ as:
% and an attribute aggregation operator $\attributeaggregationoperator$,
% we define:
% \[
% \attributeaggregatedvalue[\task]^{i}=\attributeaggregationbigoperator_{a,v\mid a\in\attributenameset\cap\attributedimension[\task]^{i},v\in\attributevalueset}\evaluationprojectionfunction{\task}{a,v}
% \]
\[
\attributeaggregatedvalue[\task]^{d}=\attributeaggregationbigoperator_{(a,v)|a\in\attributenameset}\evaluationprojectionfunction{\task}{a,v,d}
% \attributeaggregatedvalue[\task]^{i}=\attributeaggregationbigoperator_{(a,v)|a\in\attributenameset}\evaluationprojectionfunction{\task}{a,v,d}
%\attributeaggregatedvalue[\task]^{i}=\attributeaggregationbigoperator_{a,v,d\mid a\in\attributenameset,d\in\attributedimension[\task]^{i},v\in\attributevalueset}\evaluationprojectionfunction{\task}{a,v,d}
\]
\noindent be the aggregate compliance value across all task attributes
for which $d\in\attributedimension[\task]$ is the dimension relevant to an attribute $a$ in \gls{task} $t$,
% for which $d\in\attributedimension[\task]^{i}$ is the dimension relevant to an attribute $a$ in \gls{task} $t$,
$v$ is the value of the attribute $a$ in \gls{task} $t$.
In addition,
we denote $\evaluationprojectionfunction{\task}{a,v,d}=null$ if dimension $d$ does not apply to attribute $a$ in \gls{task} $t$.  Thus, any compliance dimension with a null value will be simply ignored from the aggregation.  
% for which dimension $\attributedimension[\task]^{i}$ is relevant to an attribute $a$ in \gls{task} $t$,
% $v$ is projected value of the attribute $a$ through the relation mapping function $m$ defined above,
Finally, $\attributeaggregatedvalue[\task]^{i}\in \range{0}{1}$.
%\set{0,1}$.
\end{definition}

\begin{definition}[\Glsdesc{d-compliant}]\label{definition:dCompliant}
Given a \gls{task} $\task\in\taskidset$,
$\attributedimensiondefinition[\task]$ as the set of compliance dimensions that relate to its attributes,
$\attributeaggregatedvalue[\task]^{i}$ as the aggregate attribute compliance metric per \Cref{definition:dimensionIndex},
% the set of attribute dimension $\attributedimensiondefinition[\task]$,
and $\metriccutoffvalue^{i},\metriccutoffthreshold^{i}\in\mathbf{R}$, 
then,
\begin{itemize}
\item $\task$ is \emph{non-compliant} on dimension $\attributedimension[\task]^{i}$
%  iff $\attributeaggregatedvalue[\task]^{i}=0$;
% \item $\task$ is \emph{\gls{weakly compliant}} on dimension $\attributedimension[\task]^{i}$
  iff $\attributeaggregatedvalue[\task]^{i}<\metriccutoffvalue^{i}$;
\item $\task$ is \emph{\gls{partially compliant}} on dimension $\attributedimension[\task]^{i}$
  iff $\metriccutoffvalue^{i}\leq\attributeaggregatedvalue[\task]^{i}<\metriccutoffvalue^{i}+\metriccutoffthreshold^{i}$; and
\item $\task$ is \emph{\gls{fully compliant}} on dimension $\attributedimension[\task]^{i}$
  iff $\attributeaggregatedvalue[\task]^{i}\geq\metriccutoffvalue^{i}+\metriccutoffthreshold^{i}$
\end{itemize}

\noindent where $\metriccutoffvalue[\task]^{i}$
and $\metriccutoffthreshold[\task]^{i}$ are the standard 
and threshold values
%\footnote{The cut-off and threshold values
%are numeric constants provided by analysts.}
for {full}\glsadd{full compliance}
and \gls{partial compliance}, respectively.  Note that the  threshold represents a range or window around the standard value in which partial compliance is possible.  
These values are generally numeric constants provided by the domain experts to the analysts.
\end{definition}

\Cref{definition:dimensionIndex,definition:dCompliant} define
how the attribute metric value should be calculated
and conditions for different levels of compliance,
respectively.
This means that a \gls{task} is \gls{fully compliant} if it is executing under some \emph{ideal} situation;
while a \gls{task} is \gls{partially compliant} if its
attributes in $\attributedimension[\task]$ are to a large extent in accordance with the requirements specified
but a few of them have been violated
and remedial actions have been performed to repair/compensate the situation
such that all violations identified have either been resolved or compensated;
or a task is
non-compliant otherwise.
% While \gls{weakly compliant} means
% that only some of the attributes requirements have been fulfilled
% and the majority do not, non-compliant otherwise.

% Above we have discussed how an attribute dimension metric may be assigned on each dimension of a task.
The \Gls{d-compliance} score on dimension $\attributedimension[\task]^{i}$ is given by $\attributeaggregatedvalue[\task]^{i}$
and is a real value in \range{0}{1}.
For a non-numeric value,
the attribute dimension metric may be recorded on a qualitative scale
such as a 3-point scale of  (low, medium, high)
or on a 5- or 7-point Likert scale.
In this case,
the points on the scale can be mapped uniformly to the 0-1 scale.
Thus, by default,
high would correspond to 1, medium to 0.67
and low to 0.33.
Alternatively,
a user-defined mapping function may be employed for this purpose.
In general,
rules can also be applied to determine a user-defined mapping function for nominal compliance values.
Thus,
given a task with an attribute $a$ and  a 3-point scale of (low, medium, high)
in dimension $d$,
a set of rules can be written as follows using three reasonable cut-off values:

\begin{equation*}
\evaluationprojectionfunction{}{a,v,d}=\left\{\begin{array}{r@{~~}l}
0.25 & \text{if $v$=``low''} \\[.2em]
0.50 & \text{if $v$=``medium''} \\[.2em]
0.90 & \text{if $v$=``high''}
\end{array}\right.
\end{equation*}
\smallskip

Once the individual attribute value has been evaluated,
they can be combined in different ways to obtain a dimension compliance score $\attributeaggregatedvalue[\task]^{i}$.

Below are some alternative methods to
compute the aggregation operator \attributeaggregationoperator\space for an attribute.

\begin{enumerate}
\item\emph{Average method.}
  Take a (weighted) average of attribute dimension metric values.
  This will give an average across the individual scores across all the applicable dimensions.
  For three dimensions with scores of 0.7, 0.9 and 1,
  the average would be 0.867.
  It is also possible to assign different weights to each dimension based on its importance.
\item\emph{Product method.}
  Take the product of all attribute dimension metric values.
  In this case, we would multiply across all the \evaluationprojectionfunction{}{a,v,d}'s.
  Thus, in the above example we would obtain 0.63.
  In general the product approach would lead to a lower value than the average approach.
\item\emph{Rule-based method.}
  % It is also possible to apply a more general rule-based method to combine the individual metrics.
  Apply a more general rule-based method to combine the individual metrics.
  Thus, a rule could be expressed as:
  \[
  \text{If }(\evaluationprojectionfunction{}{a_1,v_1,d_1} < 0.5)\text{~AND~}(\evaluationprojectionfunction{}{a_2,v_2,d_2} < 0.5)\text{~then~}\attributeaggregatedvalue[\task]^{i}=0.
  \]
  \noindent which states that if the partial compliance on metrics 1 and 2 is less than 0.5 then the task is non-compliant even though it is partially compliant on individual metrics, perhaps because these two metrics are very important.
\end{enumerate}

\noindent The simplest implementation of $\attributeaggregatedvalue[\task]^{i}$ % (in each dimension)
is to set $\attributeaggregationoperator$ to the (weighted) average of all non-null compliance values after evaluations,
i.e., $\attributeaggregatedvalue[\task]^{i}=\frac{1}{\sizeof{\attributedimension[\task]^{i}}}\Sigma\evaluationprojectionfunction{\task}{a}$.
However,
we should be cautious
when selecting which function to use in computing $\attributeaggregatedvalue^{i}_{t}$
as setting $\attributeaggregationoperator=max$ would mean
that whenever an attribute in a dimension is fully compliant,
then the task will also be fully compliant in this particular dimension
and similar will apply when we set $\attributeaggregationoperator=min$,
which may not be something that we intended.

\begin{example}
A review loan application task has $\metriccutoffvalue^{i}=\SI{3}{\days}$.
$\metriccutoffthreshold^{i}=\SI{2}{\days}$.
If the task takes \SI{4}{\days},
it is \gls{partially compliant} on the dimension $\attributedimension^{T}$\footnote{From now on,
we will use $\attributedimension^{M}$,
$\attributedimension^{T}$,
$\attributedimension^{R}$,
$\attributedimension^{D}$,
and $\attributedimension^{Q}$
to denote the monetary,
time,
role,
data,
and quality
dimensions of a \gls{task},
respectively.}.
But if it takes \SI{6}{\days},
it is \gls{non-compliant}.
\end{example}

Based on the definitions above,
the level of compliance of \glspl{task} can be defined in terms of a metric outside a permitted range for one
or more related dimensions,
such as money,
time, role, data, quality, etc.
% These are the typical dimensions associated with \glspl{task} of a process.
Thus,
a \gls{task} in the process may be required to be performed by a worker in a role using certain data inputs or documents.
There is also a time limit for the completion of a task
and a quality requirement.
Finally,
some \glspl{task} may also require the monetary payment of a fee
(e.g. an application fee for admission to a school,
processing fee for issuance of a passport or permit, etc.).

%If a \gls{task} satisfies all the allowed ranges for the various dimensions,
%then it is in full compliance.
%If there are deviations outside the allowed range
%but still within limits for a \gls{task} then
%it is classified as being \emph{partially compliant}.
%In such a case a compliance score,
%$\taskmeasure${-}Measure,
%is associated with each \gls{task} as a composite score obtained by combining the individual scores along the various dimensions.
% % as we shall describe shortly.
%Accordingly,
%we have the following definitions
%which characterize the status
%and level of compliance of a given \gls{task}.

\begin{definition}[\gls{t-compliance}]\label{definition:tCompliant}
Given a \gls{task} $\task\in\taskidset$, and
$\attributedimensiondefinition[\task]$ as the compliance dimensions that correspond to its various attributes, then
we define:
\begin{itemize}
\item $\task$ is \emph{non-compliant} iff $\exists\attributedimension[\task]^{i}\in\attributedimension[\task]$,
  $\attributedimension$ is \emph{\gls{non-compliant}};
\item $\task$ is \emph{\gls{fully compliant}} iff $\forall\attributedimension[\task]^{i}\in\attributedimension[\task]$,
  $\attributedimension$ is \emph{\gls{fully compliant}};
\item otherwise, $\task$ is said to be partially-compliant meaning
  that some attributes are operating under  \emph{sub-ideal} conditions.
\end{itemize}
%    and
% \item $\task$ is \emph{\gls{weakly compliant}}
%   if either $\exists\attributedimension\in\attributedimension[\task]$
%   s.t. $\attributedimension$ is \gls{partially compliant},
%   or all $\attributedimension\in\attributedimension[\task]$ are either strongly-compliant
%   or not-compliant.
% \end{itemize}
\end{definition}

% That is,
% a \gls{task} is \gls{fully compliant} iff all attribute dimension indexes are strongly compliant,
% and \gls{non-compliant} iff all attribute dimension indexes are non-compliant.
% Anything in between will be considered as \gls{sub-compliant}
% as some of the attributes may

\begin{definition}[$\taskmeasure$-Measure]\label{definition:taskComplianceMeasure}
Given a \gls{task} $\task\in\taskidset$,
$\attributedimensiondefinition[\task]$;
$\attributeaggregatedvalue[\task]^{i}$
the set of its attribute dimensions
and dimension metrics as defined in \Cref{definition:dimensionIndex};
and a dimension aggregation operator $\attributedimensionaggregationoperator$,
then we define:
\[
\pmeasure[\task]=\attributedimensionaggregationbigoperator_{i\in\range{1}{\sizeof{\attributedimension[\task]}}}\attributeaggregatedvalue[\task]^{i}
\]
\noindent as
the \gls{task compliance measure},
or $\taskmeasure$-Measure,
of \gls{task} $\task$.
\end{definition}

%Essentially,
The dimension aggregation operator $\attributedimensionaggregationoperator$ here works much like
the attribute aggregation operator $\attributeaggregationoperator$  in \Cref{definition:dCompliant}.
It aggregates dimension metrics
that were calculated for each dimension
and returns a single value that represents the overall level of task compliance.
%that can be used to indicate the level of compliance of the measured \gls{task}.
However, as discussed above, the aggregation function should be chosen with care.

% \begin{definition}[\Glsdesc{p-compliant}]\label{definition:pCompliant:task}
% An activity is \gls{p-compliant} on Metric $M_i$
% if $S_i<M_i\leq S_i+\Delta_i$,
% i.e., the metric is above the standard cut-off for compliance
% but below the cut-off after allowing for a margin of $\Delta_i$.
% \end{definition}

% \begin{definition}\label{definition:pCompliant:process}
% An process instance is \gls{p-compliant} on Metric $M_i$
% if $S_i<M_i\leq S_i+\Delta_i$,
% i.e., the metric is above the standard cut-off for compliance
% but below the cut-off after allowing for a margin of $\Delta_i$.
% \end{definition}

\begin{example}
A loan application process consists of 5 activities from submit application to receive final decision.
The standard amount of time for it is \SI{15}{\days}.
If the threshold $\Delta$ is \SI{5}{\days}
and it takes \SI{18}{\days} to finish the loan application process,
then it is \gls{partially compliant},
showing that even when some activity(ies) in the process instance may be \gls{non-compliant},
% The $\Delta$ is \SI{5}{\days}.
% If it takes \SI{18}{\days} it is \gls{p-compliant}.
% Note that some activity(ies) in the process instance may be \gls{non-compliant},
the instance itself can be compliant.
\end{example}

Consequently,
given an instance of \gls{trace} $\trace$ of a \gls{bp},
one can simply calculate the level of compliance of $\trace$
by directly aggregating/averaging the $\taskmeasure$-Measure value of each \gls{task}.
However,
this may have some drawbacks as the aggregated value
may not necessarily reflect the real situation of the \emph{whole} \gls{trace}.
This is due to the fact
that the changes made after any non-compliance issues might introduce
new attributes (and/or values),
and changes to the task. Besides,
during execution other tasks may also impact
the value of the attribute, averaging
these values might not
give correct performance of the
attribute, hence it would not make sense.
% \textcolor{red}{This is due to the fact
% that \glspl{task}
% that appear at the beginning of the \gls{trace} normally perform better,
% or without any compliant issues;
% while \glspl{task} at the later part of the \gls{trace} not.
% Besides, the value of the same attribute may have changed by different \glspl{task} during execution,
% averaging these values actually concealed the real performance of the attribute
% and does not make sense.}

To overcome these issues,
we define  \gls{trace} compliance
and a \gls{trace} compliance measure based on the attribute dimension metrics,
as follow.

\begin{definition}[\trace-compliance]\label{definition:traceCompliant}
Given an instance of \gls{trace} $\trace$ of a \gls{bp},
we define:
\begin{itemize}
\item $\trace$ is \emph{\gls{non-compliant}}
  iff $\exists\task\in\taskidset$, $\task$ is \emph{\gls{non-compliant}};
\item $\trace$ is \emph{\gls{fully compliant}}
  iff $\forall\task\in\taskidset$, $\task$ is \emph{\gls{fully compliant}};
\item otherwise,
  $\trace$ is said to be partially-compliant meaning
  that $\trace$ has been executed under some \emph{sub-ideal}
  (or \emph{sub-optimal}) conditions.
\end{itemize}
\end{definition}

\begin{definition}[$\tracemeasure$-Measure]\label{definition:traceMeasure}
Given an instance of \gls{trace} $\trace$ of a \gls{bp};
$\taskidset$ the set of unique task identifiers of tasks;
$\attributedimensiondefinition[\trace]$;
$\attributeaggregatedvalue[\task]^{i}$
the set of attribute compliance dimensions
that appear in $\trace$;
the aggregate compliance  metric of task $\task$ as in \Cref{definition:dimensionIndex};
and $\pmeasureaggregationoperator$ the \gls{task} dimension aggregation operator,
then we define the trace partial compliance measure as:
% \[
% \pmeasure[\trace]=\pmeasureaggregationbigoperator_{\attributedimension[\trace]^{i}\mid\attributedimension[\trace]^{i}\in\attributedimension[\trace]}\,\attributelastupdateoperator(\attributeaggregatedvalue[\trace]^{i})
% \]
\[
\pmeasure[\trace]=\pmeasureaggregationbigoperator_{\attributedimension[\trace]^{i}\mid\attributedimension[\trace]^{i}\in\attributedimension[\trace]}\,\argmin_{\task\mid\task\in\taskidset\cap\attributedimension[\trace]^{i}\in\attributedimension[\task]}(\attributeaggregatedvalue[\task]^{i}>0)
\]
% be the \gls{trace}-compliance measure,
% or $\tracemeasure$-Measure,
% of the trace \trace.
% where $\attributelastupdateoperator(\attributeaggregatedvalue[\trace]^i)$ is the \emph{last updated} metric value of dimension $\attributedimension[\trace]^{i}$ in trace \trace.
\end{definition}

As execution progresses,
the aggregate compliance  metrics of each \gls{task} will be updated accordingly.
% In most cases,
% to reflect the situations
% when a violation appear,
% these values will be reduced (normally).
Hence,
to reflect this situation,
% here,
the compliance measure of a \gls{trace} is defined by the aggregated dimension metrics (across
all dimensions).
%that is either fully or \gls{partially compliant}.
Naturally, if all metrics of a particular dimension are $0$ for a task, % \glspl{task},
then a zero value will be returned.
% that appear in the \gls{trace}.
% Accordingly,
% for a \gls{trace} with a \gls{task}
% that cannot be accomplished,
% i.e., certain of its dimension metrics will become zero,
% indicating that the \gls{trace} is (totally) non-compliant.
% Also
Note here
that an instance of \gls{trace} can be \gls{d-compliant} on multiple dimensions, yet
it does not mean that it will automatically be $\trace$-compliant at the end.

% \begin{example}
% The review loan application takes \SI{4}{\days},
% it costs \SI{70}[\$]{}
% which is above the standard cost of \SI{50}[\$]{},
% and out of the $5$ documents required
% it is missing one document,
% say,
% the client's bank account statement.
% % \cq{something else\ldots?}
% In this case, the task will be incomplete and
% the trace will not be properly executed,
% leading to temporary suspension of the process.
% \end{example}

Lastly,
we give the following definition for the overall compliance of a process log consisting of multiple traces to conclude our framework.

\begin{definition}[\processmeasure-Measure]
Given a \gls{bp} $\process$;
$\traceinstanceset[\process]$ the set of log \gls{trace} instances obtained after executing \process;
and $\sizeof{\traceinstanceset[\process]}$, its size,
then the compliance measure for the process $\process$
is given by:
\[
\processmeasure=\frac{1}{\sizeof{\traceinstanceset[\process]}}\Sigma_{\trace\in\traceinstanceset[\process]}\tracemeasure
\]
\noindent where $\tracemeasure$ is the \tracemeasure-Measure of the \gls{trace} instance \trace.
\end{definition}

Here,
the compliance measure of a \gls{bp},
\processmeasure-Measure,
is defined as the average  value of the
\tracemeasure-Measure across all \glspl{trace} since
%,unlike the situation that appear in \glspl{task},
each \gls{trace} represents an independent execution of $\process$
and will not affect other ones.

It is important to note that we have defined our metrics at three levels of aggregation in a hierarchical manner, i.e., at the task, trace and process log levels.  
Depending upon the user application and requirements, metrics at one or more levels can used in conjunction with each other to gain multiple perspectives.  
% It is also possible
Besides,
it is possible 
that a metric may be violated at one level
but may still be satisfied at another or a higher level, or vice-versa. 
Moreover, some metrics along some dimensions like time may be more meaningful at the instance level as in Example 2 since the total instance duration is more important for the customer than the duration of individual tasks.  Other metrics may be more relevant at the task level, such as the monetary amounts involved, etc.  The process log level metrics can give insights into the overall compliance level for the entire log over a period of time, such as a week, month or quarter. Comparing such metrics across several successive periods can provide managerial insights into  overall compliance trends.

\section{Composite Measure Computations}
\label{section:exampleOfMeasureCalculation}

% \q{\Cref{table:attributesMetric} needs to be updated with better metric values}

% \begin{table}\centering
% \caption{Attribute values w.r.t. different \glspl{task}}
% \label{table:attributeValuesInTasks}
% {
% \newcolumntype{Y}{>{\centering\arraybackslash}X}
% \begin{tabularx}{.7\linewidth}{lYYYY}
% \toprule
% & \task[2] & \task[3] & \task[4] & \task[5] \\
% \midrule
% paymentReceive & $\leq$15 & $\leq$22 & $\leq$32 & $>$32 \\
% \bottomrule
% \end{tabularx}
% }
% \end{table}

% \toupdate{
% \begin{itemize}
% \item Define aggregation operators to be used in the running example:
%   \attributeaggregationoperator (\Cref{definition:dimensionIndex}),
%   \attributedimensionaggregationoperator (\Cref{definition:taskComplianceMeasure}),
%   \pmeasureaggregationoperator (\Cref{definition:traceMeasure})
% \item scenarios: pay on time, with 1 default, and with 3 defaults (contract termination)
% \end{itemize}
% }

%In this section,
Next, we discuss  some scenarios in  the context of a real-world example to illustrate
how the proposed framework can be applied in practice
to compute
different levels of compliance by employing the
\emph{averaging method} discussed in the previous section.
For this purpose, we
consider the invoice payment
example from \Cref{figure:fragmentOfPaymentProcessModel},
and provide some notation for our computations.
We consider the attribute aggregation operator $\attributeaggregationoperator$
to be the average of all attribute values projected in the dimension,
i.e.,
$\attributeaggregatedvalue[\task]^{i}=\frac{1}{\sizeof{\attributedimension[\task]^{i}}}\Sigma\evaluationprojectionfunction{\task}{a}$.
Moreover,
$\attributedimensionaggregationoperator$ is
the compliance dimension aggregation operator averaging
the dimension index values for each dimension in the task i.e.,
% $\attributedimensionaggregationbigoperator_{i\in\set{1,\sizeof{\attributedimension[\task]}}}\attributeaggregatedvalue[\task]^{i}$
$\attributeaggregatedvalue[\task]^{i}=\frac{1}{\sizeof{\attributedimension[\task]^{i}}}\Sigma\evaluationprojectionfunction{\task}{a}$,
and $\pmeasureaggregationoperator$
is the minimum of all values for each
dimension.
%appearing in the trace s.t. $\attributeaggregatedvalue[\task]^{i}=\argmin$.

\Cref{table:attributesMetric} illustrates the attributes and their possible values
in the context of  \Cref{figure:fragmentOfPaymentProcessModel}.
Attributes such as $description$, $invoiceValue$ and $invoiceDate$
are meta information of the invoice
and do not contribute to the compliance metrics.
The attributes $equipmentDeliveryDays$
and $payInDays$ denote the number of days required to deliver the equipment(s) to the purchaser
and the number of days
% The attribute $payInDays$ denotes the number of days
within which \emph{full} payment  must be made after the invoice is issued, respectively.
As shown, different values for these parameters are mapped to
compliance levels based on the $\evaluationprojectionfunctionsimple$ projection function.
Moreover, in this scenario,  the partial compliance cut-off value $\metriccutoffvalue$
and the threshold $\metriccutoffthreshold$ are set to $0.3$ and $0.4$,
respectively.  Thus,
 a compliance value between $0.3$
and $0.7$ $(0.3+0.4)$ is considered as \emph{\gls{partially compliant}},
and any value below 0.3  as \emph{\gls{non-compliant}}.
Similarly, the attribute $paymentReceived$ is  the amount
 paid by the customer,
which includes the principal  plus any applicable interest and penalty.
%that may possibly appear.
% Notice here
% that we have set the $\metriccutoffvalue$
% and $\metriccutoffthreshold$ to $0.3$
% and $0.7$,
% respectively,
% indicating that only payment received
% that is greater than or equal to the payable amount will be considered as strongly compliant.
Notice from the table that a payment
 of less than  half of the  amount due is deemed as non-compliant,
while other values of payment are considered as partially compliant.
The two attributes, $interest$ and $penalty$
are meta information that will be used to calculate the penalty when violations occur.
%In addition,
%we also provide the partial compliance for different values of $\evaluationprojectionfunction{\task[6]}{equipmentDeliveryDays}$
%in \Cref{table:attributesMetric}
%for the case when $payInDays>0$,
%or $0$ otherwise.

\begin{table*}\centering
\caption{Attributes metric of \glsname{bp} in \Cref{figure:fragmentOfPaymentProcessModel}}
\label{table:attributesMetric}
% \scriptsize
\footnotesize
%!TEX root = ../manuscript.tex

{
\begin{tabular}{l@{~}ccc@{~~}c@{~~}c}
\toprule
\multicolumn{1}{>{\bfseries}c@{\quad}}{\Gls{attribute} name ($a$)}
  & \textbf{Dimension}
  & \textbf{\attributenameset}
  & \textbf{\evaluationprojectionfunction{}{a}}
  & \textbf{Cut-off ($\metriccutoffvalue$)}
  & \textbf{Threshold ($\metriccutoffthreshold$)}
  \\
% \midrule
% $description$ & -- & Make payment & -- & -- & -- \\
\midrule
$invoiceValue~(P)$ & Monetary & \SI{500}[\$]{} & -- & -- & -- \\
\midrule
$invoiceDate$ & Temporal & 2019-04-01 & -- & -- & -- \\
\midrule
\multirow{3}{*}{$equipmentDeliveryDays$}
  & \multirow{3}{*}{Temporal}
  & $\leq$\SI{3}{\days} & $1$
  & \multirow{7}{*}{$0.3$}
  & \multirow{7}{*}{$0.4$}
  \\
&& $\leq$\SI{7}{\days} & $0.5$ \\
% && $\leq$\SI{10}{\days} & $0.1$ \\
&& $>$\SI{7}{\days} & $0$ \\
\cmidrule{1-4}
\multirow{4}{*}{$payInDays$}
  & \multirow{4}{*}{Temporal}
  & $\leq$\SI{15}{\days} & $1$
  % & \multirow{4}{*}{$0.3$}
  % & \multirow{4}{*}{$0.4$}
  % & \multirow{4}{*}{$0.3$}
  % & \multirow{4}{*}{$0.4$}
  \\
&& $\leq$\SI{22}{\days} & $0.6$ \\
&& $\leq$\SI{32}{\days} & $0.3$ \\
&& $>$\SI{32}{\days} & $0$ \\
\midrule
\multirow{2}{*}{$interest~(Int)$}
  & \multirow{2}{*}{Percentage}
  & \SI{0}{\percent} & \multirow{2}{*}{--} & \multirow{2}{*}{--} & \multirow{2}{*}{--}
  \\
&& \SI{3}{\percent} & \\
\midrule
\multirow{2}{*}{$penalty~(Pen)$}
  & \multirow{2}{*}{Percentage}
  & $0$ & \multirow{2}{*}{--} & \multirow{2}{*}{--} & \multirow{2}{*}{--}
  \\
&& \SI{2.5}{\percent} & \\
\midrule
\multirow{5}{*}{\begin{varwidth}{\linewidth}$paymentReceived$\\ $(R=P+Int+Pen)$\end{varwidth}}
  & \multirow{5}{*}{Monetary}
  & $<\SI{50}{\percent}\times R$
  % & \SI{250}[\$]{}$<$ (\SI{50}{\percent}) 
  & 0
  & \multirow{5}{*}{0.3}
  & \multirow{5}{*}{0.7} \\
&& $<\SI{75}{\percent}\times R$ & 0.3 \\
&& $<\SI{80}{\percent}\times R$ & 0.5 \\
&& $ <R$ & 0.9 \\
&& $\geq$R & 1 \\
% \midrule
% \multicolumn{1}{c}{\vdots}
%   & \multicolumn{1}{c}{\vdots}
%   & \vdots
%   & \vdots
%   & \vdots
%   & \vdots
%   \\[.2em]
% $\attribute[i][n]$ & $S^{n}_i$ & $\Delta^{n}_i$ \\
\bottomrule
% \end{tabularx}
\end{tabular}
}

% \medskip
\end{table*}

\begin{description}
%Consider now the scenario
\item[Full Compliance:]
% when $payInDays=\SI{10}{\days}$
Consider a scenario
where
$equipmentDeliveryDays=\SI{2}{\days}$,
$payInDays=\SI{10}{\days}$
and a payment of \SI{500}[\$]{} has been received from the purchaser,
i.e., the equipment has been delivered
and full payment has been received within the prescribed time frame.
Hence, as an example,
consider the trace
$\trace[4]=\angleset{\transition[1],\transition[2],\transition[6]}$
which contains the attributes $\langle paymentReceived,payInDays$, $equipmentDeliveryDays \rangle$,
as illustrated in
\Cref{table:attributesMetric}. To compute the aggregate
metric  across the compliance dimensions for an attribute,
we first compute the individual compliance values along each dimension and then aggregate them.
The compliance metric for the monetary (M) and temporal (T) dimensions for \gls{task} $\task[2]$ are first computed as:
%\footnote{Here, we abbreviated ``Monetary''
%and ``Temporal'' as ``M'' and ``T'', respectively.
%Besides, we have assumed that \task[1] and \task[5] do not contain any attribute.}
%:
\medskip

\begin{center}\vspace*{-.8em}
{\renewcommand{\arraystretch}{\tabulararraystretch}
\begin{tabular}{>{$}l<{$}@{~=~}>{$}l<{$}}
\attributeaggregatedvalue[\task_2]^{M}
  & \frac{1}{\sizeof{\attributedimension[\task_2]^{M}}}\Sigma_{a\in\attributedimension[\task_2]^{M}}\evaluationprojectionfunction{\task_2}{a}
  \\
& \frac{1}{\sizeof{\attributedimension[\task_2]^{M}}}(\evaluationprojectionfunction{\task_2}{paymentReceived})
  \\
& \frac{1}{1}(1)
  \\
& 1
  % \\[.7em]
\end{tabular}
}
{\renewcommand{\arraystretch}{\tabulararraystretch}
\begin{tabular}{>{$}l<{$}@{~=~}>{$}l<{$}}
\attributeaggregatedvalue[\task_2]^{T}
  & \frac{1}{\sizeof{\attributedimension[\task_2]^{T}}}\Sigma_{a\in\attributedimension[\task_2]^{T}}\evaluationprojectionfunction{\task_2}{a}
  \\
& \frac{1}{\sizeof{\attributedimension[\task_2]^{T}}}(\evaluationprojectionfunction{\task_2}{payInDays})
  \\
& \frac{1}{1}(1)
  \\
& 1
\end{tabular}
}
\end{center}

% $\attributeaggregatedvalue[\task_2]^{M}=\frac{1}{\sizeof{\attributedimension[\task_2]^{M}}}\Sigma_{a\in\attributedimension[\task_2]^{M}}\evaluationprojectionfunction{\task_2}{a}=\frac{1}{\sizeof{\attributedimension[\task_2]^{M}}}(\evaluationprojectionfunction{\task_2}{paymentReceived})=\frac{1}{1}(1)=1$
% \smallskip

% $\attributeaggregatedvalue[\task_2]^{T}=\frac{1}{\sizeof{\attributedimension[\task_2]^{T}}}\Sigma_{a\in\attributedimension[\task_2]^{T}}\evaluationprojectionfunction{\task_2}{a}=\frac{1}{\sizeof{\attributedimension[\task_2]^{T}}}(\evaluationprojectionfunction{\task_2}{payInDays})=\frac{1}{1}(1)=1$

% % $\attributeaggregatedvalue[\task_6]^{T}=\frac{1}{\sizeof{\attributedimension[\task_6]^{T}}}\Sigma_{a\in\attributedimension[\task_6]^{T}}\evaluationprojectionfunction{\task_6}{a}=\frac{1}{\sizeof{\attributedimension[\task_6]^{T}}}(\evaluationprojectionfunction{\task_6}{equipmentDeliveryDays})=\frac{1}{1}(1)=1$
% \smallskip

\noindent Further, by \Cref{definition:taskComplianceMeasure}, we have:

{\renewcommand{\arraystretch}{\tabulararraystretch}
\begin{tabular}{>{$}l<{$}@{~=~}>{$}l<{$}}
\taskmeasure[\task_2]\text{-Measure}
 & \frac{1}{\sizeof{\attributedimension[\task_2]}}\Sigma_{i\in\attributedimension[\task_2]}\attributeaggregatedvalue[\task_2]^{i}
 \\
& \frac{1}{\sizeof{\attributedimension[\task_2]}}(\attributeaggregatedvalue[\task_2]^{T}+\attributeaggregatedvalue[\task_2]^{M})=\frac{1}{2}(1+1)
  \\
& 1
\end{tabular}
}

% \vspace*{-1.5em}
% \[
% \taskmeasure[\task_2]\text{-Measure}=\frac{1}{\sizeof{\attributedimension[\task_2]}}\Sigma_{i\in\attributedimension[\task_2]}\attributeaggregatedvalue[\task_2]^{i}=\frac{1}{\sizeof{\attributedimension[\task_2]}}(\attributeaggregatedvalue[\task_2]^{T}+\attributeaggregatedvalue[\task_2]^{M})=\frac{1}{2}(1+1)=1
% \]

% % \smallskip

\noindent Similarly, for \gls{task} \task[6], we have: $\attributeaggregatedvalue[\task_6]^{T}=1$
and $\taskmeasure[\task_6]\text{-Measure}=1$.

\smallskip

% The projected value for attributes at monetary and temporal dimensions for
% $\transition[2] \mbox{and} \transition[6]$ is equal to $1$.
% % $\evaluationprojectionfunction{\task_2}{paymentReceived}=1$ and
% % $\evaluationprojectionfunction{\task_2}{payInDays}=0.6$, respectively.
% We then compute the average metric score by aggregating the
% attribute index values for each dimension of each task.
% \[
%  \begin{array}{l@{~=~}l@{~=~}l}
%  \taskmeasure[\task_2]\text{-Measure} & \frac{1}{2}(1 + 1) & 1\\[.5em]
%  \taskmeasure[\task_6]\text{-Measure} & \frac{1}{1}(1) & 1\\
% \end{array}
% \]

% \[
% \taskmeasure[\task_6]\text{-Measure}=\frac{1}{\sizeof{\attributedimension[\task]}}\Sigma_{i\in\attributedimension[\task]}\attributeaggregatedvalue[\task]^{i}=\frac{1}{\sizeof{\attributedimension[\task]}}(\attributeaggregatedvalue[\task_6]^{T})=\frac{1}{1}(1)=1
% \]

% Then, we compute $\taskmeasure[\task_2]$-Measure$=0.8$
% as average metric score
% indicating the compliance measure of aggregated dimension
% index values computed of the measured task.

% Finally, we compute the trace compliance
% measure as the average score of
% all $\taskmeasure[\trace_1]$-\mbox{Measure} values for each task of the trace.
Hence, we have:
$\attributedimension[\trace_1]=\set{\attributedimension[\trace_1]^{M},\attributedimension[\trace_1]^{T}}$,
and
\mbox{$\argmin({\attributedimension[\trace_1]^{M}})=1$},
and $\argmin({\attributedimension[\trace_1]^{T}})=1$.

% \[
%  \begin{array}{l@{~=~}l@{~=~}l}
Consequently, it follows that: $\taskmeasure[\trace_1]\text{-Measure}=\frac{1}{2}(1 + 1) =1$.
%    \end{array}
% \]

% Notice, $\task[1] \mbox{and} \task[6]$
% are compliant by default as they
% do not affect the computations
% for task $\task[2]$ in the scenario.
\medskip
\item[Partial Compliance:]\fussy
% Now,
% consider the situation
% when $payInDays=\SI{20}{\days}$
% and a payment of \SI{380}[\$]{} (\SI{76}{\percent}) has been received
Let us now turn to consider  a different  scenario where:\\
$equipmentDeliveryDays=\SI{2}{\days}$, \\
$payInDays=\SI{20}{\days}$\\
%$paymentReceived=\SI{\$575}$\\
$paymentReceived=\SI{575}[\$]{}$\\
%and a payment of \SI{575}[\$]{} have been received.
\noindent In this scenario,
the payable amount is now
$\SI{500}[\$]{}+\SI{75}[\$]{}=\SI{575}[\$]{}$,
% $ 500 + 75 = 575$,
and has been fully paid by the customer.Thus, we can calculate the aggregate partial compliance measures  for the tasks as follows:

{\renewcommand{\arraystretch}{\tabulararraystretch}
\begin{tabular}{>{$}l<{$}@{~=~}>{$}l<{$}}
interest
  & (20-15)\times\SI{0.03}{\percent} \times \SI{500}[\$]{}
  \\
& \SI{75}[\$]{}
  \\
\multicolumn{2}{l}{\text{ and }$penalty = 0$.}
\end{tabular}
}

% % \vspace*{-1em}
% \[
% \begin{array}{l@{~=~}l@{~=~}l}
% interest & (20-15)\times\SI{0.03}{\percent} \times \SI{500}[\$]{} & \SI{75}[\$]{} \\
%   % & 0.15 \times\SI{500}[\$]{}  \\
%   % & \SI{75}[\$]{} \\
% \end{array}\qquad\text{ and }\qquad penalty = 0
% \]

% Even though a payment of $\SI{500}[\$]{}$ has been received,
% which is equal to the original payable value,
% it is only equivalent to $\SI{86.96}{\percent}$ of the current amount required.
Accordingly, we have the following:

% \begin{tabular}[t]{cc}
% \multicolumn{2}{c}{$\task[2]$} \\
% \toprule
% Attribute (a) & $\evaluationprojectionfunctionsimple(a)$ \\
% \midrule
% payInDays & 0 \\
% paymentReceived & 0 \\
% \midrule
% $\taskmeasure[\task_2]$-Measure & 0 \\
% \bottomrule
% \end{tabular}~
% \begin{tabular}[t]{cc}
% \multicolumn{2}{c}{$\task[3]$} \\
% \toprule
% Attribute (a) & $\evaluationprojectionfunctionsimple(a)$ \\
% \midrule
% payInDays & 0.6 \\
% paymentReceived & 1 \\
% \midrule
% $\taskmeasure[\task_3]$-Measure & 0.8 \\
% \bottomrule
% \end{tabular}~
% \begin{tabular}[t]{cc}
% \multicolumn{2}{c}{$\task[6]$} \\
% \toprule
% Attribute (a) & $\evaluationprojectionfunctionsimple(a)$ \\
% \midrule
% equipmentDeliveryDay & 1 \\
% \midrule
% $\taskmeasure[\task_6]$-Measure & 0 \\
% \bottomrule
% \end{tabular}

{\begin{center}\footnotesize
\noindent\begin{tabular}{l*{4}{@{\quad}c}}
\toprule
Dimensions & Attributes & $\task[2]$ & $\task[3]$ & $\task[6]$\\
\midrule
\multirow{2}{*}{Temporal} & $equipmentDeliveryDays$ &  -- & -- & 1 \\
& $payInDays$ & 0 & 0.6 & -- \\
\midrule
Monetary & $paymentReceived$ & 0 & 1 & -- \\
\midrule\midrule
\multicolumn{2}{r}{$\taskmeasure$-Measure\quad\mbox{}} & 0 & 0.8 & 1 \\
\bottomrule
\end{tabular}
\end{center}}

Hence,
$\argmin({\attributedimension[\trace]^{M}})=0.6$,
and $\argmin({\attributedimension[\trace]^{T}})=1$.

Therefore, $\tracemeasure\text{-Measure}=\frac{1}{2}(0.6+1)=0.8$.
\\
\item[Non-Compliance]
Lastly,
consider the situation
where no payment has been received after \SI{32}{\days},
% \noindent Consequently,
% if the purchaser does not make any payment after \SI{32}{\days},
i.e., the conditions of the contract have been violated
and cannot be repaired. Thus, the contract  will be deemed as terminated.

% However, if further payment,
% which includes the interest and penalty,
% has been made by the purchaser,
% then \ldots

% Lastly,
% consider the scenario
% where $equipmentDeliveryDays=\SI{2}{\days}$
% and no payment has been received after \SI{32}{\days}.

{\begin{center}\footnotesize
\noindent\begin{tabular}{l*{5}{@{\quad}c}}
\toprule
Dimensions & Attributes & $\task[2]$ & $\task[3]$ & $\task[4]$ & $\task[6]$ \\
\midrule
\multirow{2}{*}{Temporal} & $equipmentDeliveryDays$ &  -- & -- & -- & 0 \\
& $payInDays$ & 0 & 0 & 0 & -- \\
\midrule
Monetary & $paymentReceived$ & 0 & 0 & 0 & -- \\
\midrule\midrule
\multicolumn{2}{r}{$\taskmeasure$-Measure\quad\mbox{}} & 0 & 0 & 0 & 0 \\
\bottomrule
\end{tabular}
\end{center}}

$\tracemeasure\text{-Measure}=\frac{1}{2}(0+0)=0$.

The \gls{partial compliance} functions can also be introduced as mappings from a numeric domain of attribute values corresponding to the threshold window around the standard value for an attribute to the \range{0}{1} range.  These functions typically take a linear, concave or convex form depending upon how rapidly the distance from the standard value affects compliance.  The shapes of typical functions have to be determined through empirical studies and this is out of the scope of the current work. 
% As $payInDays>\SI{32}{\days}$
% and no payment has been received,
% i.e., the conditions of the contract has been violated
% and will be terminated accordingly.

\end{description}

% Besides,
% we also assume
% that either payment has been received in full,
% or no payment has been received.
% The case of partial payment will not be consider in the discussion below.
%While we have examined different levels
%of compliance, the ideal case scenario is always desirable;
%however sub-ideal cases (partially compliant) still
%provide the possibility to execute the process.
%Essentially, partial compliance is defined in terms
%of the deviation of a metric outside an allowed range
%as demonstrated above for one or more dimensions
%related to money, time, data etc.

% \toupdate{
% Ways to overcome \gls{p-compliant}:
% compensations, repair, penalty, undo, redo, restart, abort.

% \noindent Note: we can discuss briefly each of these cases
% or as many as possible in the context of the running example.
% }

%!TEX root = ../manuscript.tex

\section{Discussion and related work}
\label{section:relatedWork}

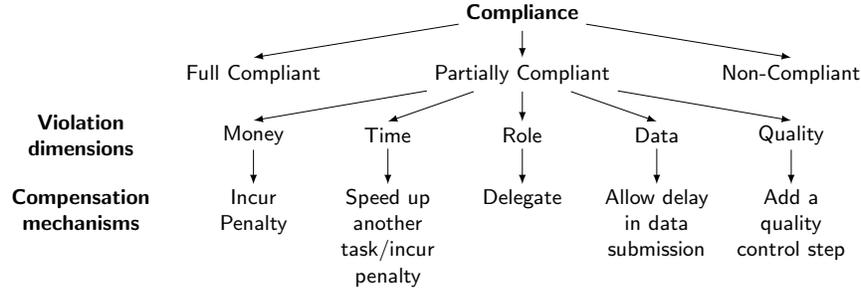
\begin{figure*}[t]\centering
\noindent\resizebox{.98\linewidth}{!}{%!TEX root = ../manuscript.tex

{
\begin{tikzpicture}[
font=\sffamily\footnotesize,
every node/.style={inner sep=3pt},
dNode/.style={text centered},
catNode/.style={dNode,text width=7em,font=\sffamily\footnotesize\bfseries},
vNode/.style={dNode,anchor=north,text width=5em},
every path/.style={shorten >=0pt,shorten <=0pt},
]

\matrix [
% column sep=1.8em,
row sep=.5em,
nodes={inner sep=3pt},
% column 1/.style={nodes={dNode,font=\sffamily\footnotesize\bfseries}},
row 1/.style={nodes={dNode}},
row 3/.style={nodes={text width=5em,text centered,minimum height=1.2em,inner sep=0pt,inner ysep=2pt}},
row 4/.style={nodes={text width=5.5em,text centered,inner ysep=2pt,anchor=north}},
% row 2/.style={nodes={vNode,text width=10em}},
] {
&[1em]&& \node (compliance) [font=\sffamily\bfseries] {{Compliance}} ;
  \\[1.2em]
&&& \node (partialLoc) {};
  \\[.8em]
\node [catNode] {Violation dimensions} ;
  & \node (dim1) {Money} ;
  & \node (dim2) {Time} ;
  & \node (dim3) {Role} ;
  & \node (dim4) {Data} ;
  & \node (dim5) {Quality} ;
  \\[.6em]
\node [catNode] {Compensation mechanisms} ;
  & \node (d1) {Incur Penalty} ;
  & \node (d2) {Speed up another task/incur penalty} ;
  & \node (d3) {Delegate} ;
  & \node (d4) {Allow delay in data submission} ;
  & \node (d5) {Add a quality control step} ;
  \\
} ;

% \node (v2) [vNode] at (v1.north-|d2) {} ;
% \node (v3) [vNode] at (v1.north-|d3) {} ;
% \node (v4) [vNode] at (v1.north-|d4) {} ;
% \node (v5) [vNode] at (v1.north-|d5) {} ;

\node (partial) at (partialLoc) {Partially Compliant} ;
\node (full) at (partial-|dim1) {Full Compliant} ;
\node (non) at (partial-|dim5) {Non-Compliant} ;

\draw [->] (compliance) -- (full.north) ;
\draw [->] (compliance) -- (partial.north) ;
\draw [->] (compliance) -- (non.north) ;

\foreach \n/\l in {1/-2.5pt,2/-1pt,3/0pt,4/1pt,5/2.5pt} {
  \draw [->] (partial) -- (dim\n.north) ;
  % \draw [->,shorten >=0pt,shorten <=0pt] ($(partial.south)+(3*\l,0)$) -- (dim\n.north) ;
  \draw [->] (dim\n) -- (d\n) ;
}

\end{tikzpicture}
}}
% \caption{Compliance dimensions and compensation mechanisms for them}
\caption{Compliance dimensions and compensation mechanisms for \gls{partial compliance}}
\label{figure:complianceDimensionsCompensationMechanisms}
\end{figure*}

There are different ways to overcome  various partial
compliance scenarios as shown in \Cref{figure:complianceDimensionsCompensationMechanisms}.
For each kind of deviation or case of partial compliance,
one or more compensatory mechanisms may be provided
for the task to resume execution.

For example, if a task is delayed it may be made up
by speeding up a later task so that the customer of a
service does not notice any increase in the total time
for a process instance. 
A role violation occurs when an employee in the designated role is not available to perform a task. In such a case, a possible compensation is to assign the task to a delegate of the person who would normally perform it.  
For the data dimension, to process
a passport application a user may be required
to provide social security card or ID card and
birth certificate, etc. If the user does not provide the
birth certificate (or, say, one of three required documents), it may still be possible
to process the application provided the missing document
is submitted within one week of the application. Thus, the application may still be processed despite a minor violation.  In the absence of such a mechanism, the application would have to be rejected, and then have to be resubmitted thus increasing the overall cost of processing it both for the citizen and for the governmental agency involved.

On the other hand, if the process instance itself
takes longer than the standard time, then the
customer has to be compensated by the service provider
as per their agreement. With regard to the process
model redo, restart, undo or abort the tasks can be
other possible ways to overcome
partial compliance issues. As they have their own
complexities and impact on the execution of the individual
tasks and the process on the whole,
these are the topics of our further investigation.

Our framework determines partial compliance in a bottom up manner from the individual task level, to the trace level, and then a compliance score can thus be assigned to each trace.  It is also possible to consider partial compliance at a still higher level of all process instances or cases in a day, week or month. The notion here would be to determine how many traces fall within a certain partial compliance level, say X\% of instances have a compliance level more than 0.8 during  a month. Moreover, a similar analysis may be done at the individual task level to determine what percentage of payment tasks had a compliance level of more than, say,  0.9 in a month. Such information can be very helpful to the management of a company. As we noted above, binary notions of compliance are not very useful from a management perspective for understanding the operational performance of a company. By introducing partial compliance in this way management can gain deeper insights into their operations.
The values in \Cref{table:attributesMetric}  can be derived through empirical studies
and from analysis of logs from previous executions of the business process model.

The problem of \gls{partial compliance} has been studied widely in different domains.
Gerber and Green \cite{Gerber.2012} proposed the use of regression analysis
scalable protocols to resolve some of the \gls{partial compliance} issues
that appear in field experiments.
Jin and Rubin~\cite{Jin.2008},
on the other hand,
proposed the use of principal stratification to handle \gls{partial compliance} issues
when analysing drug trials
and educational testing.
However,
only a limited amount of work has targeted the area of improving \gls{bpc} management
or measuring the level of compliance of a \gls{bp} in a quantitative way.
In the followings,
we present some pertinent studies
and discuss their strengths and limitations for the measurement of \gls{partial compliance}.

% \begin{figure}\centering
% \input{images/smartContractTriggerComponent}
% \caption{??}
% \end{figure}

In~\cite{Sadiq.2007a},
\etal{Sadiq} introduced the notion of compliance distance as a quantitative measure
of how much a process model may have to change in response to a set of rules (compliance objectives) at design time;
or by counting the number of recoverable violations,
how much an instance deviates from its expected behavior at runtime.
This approach is extended in~\cite{Lu.2008a} to effectively measure the distance
between compliance rules
and organization's processes.
To this purpose,
the authors have divided the control objectives into four distinct classes of ideal semantics,
namely: \begin{enumerate*}[(i)]
\item \emph{ideal},
\item \emph{sub-ideal},
\item \emph{non-compliant}, and
\item \emph{irrelevant},
\end{enumerate*}
and compute the degree to which a \gls{bp} supports the compliance rules.
Although their method provides computationally efficient means to analyze the relationships
between the compliance rules
and \glspl{bp},
the heavily formalized rules have increased the complexity of the modelling process
which is a potential obstacle to non-technical users.

\ifspringer
Shamsaei~\cite{Shamsaei.2011b}
\else
\citet{Shamsaei.2011b}
\fi
proposed a goal-oriented, model-based framework
for measuring the level of compliance of a \gls{bp} against regulations.
In the paper,
the author decomposed the regulations into different control rule levels,
and then defined a set of \glspl{kpi}
and attributes for each rule to measure their level of compliance.
The value of the \glspl{kpi} can be provided either manually
or from external data sources,
and the satisfaction level of each rule is evaluated on a scale between $0$
and $100$ by considering the values of target,
threshold,
and worst,
so that analysts can prioritize compliance issues  to address suitably given the limited resources at their disposal.

\ifspringer
\etal{Morrison}~\cite{Morrison.2009}
\else
\citet{Morrison.2009}
\fi
% \etal{Morrison}~\cite{Morrison.2009} 
proposed a generic compliance framework
to measure the levels of compliance using \emph{\gls{c-semiring}}~\cite{Bistarelli.2004a}.
In their approach,
imprecise
or non-crisp compliance requirements will first be quantified by means of \emph{decision lattices}
(through the notion of \emph{lattice chain}),
which provides a formal setting to represent concept hierarchies
and values preferences~\cite{Huchard.2007},
such as \{$Good$, $Ok$, ${Bad}$\}
or \{$Good$, $Fair$, $Bad$\}.
These values will then be combined
and utilized as a decision-making tool by \glspl{c-semiring} to rank the level of compliance of the \glspl{bp}.
Essentially,
the advantage of their framework lies in the ability in combining compliance assessments on various dimensions.
Although the proposed approach is general enough to provide an abstract valuation at policy (business process in this case) level, the
information about  compliance  at lower levels of abstraction is missing.

Kumar and Barton \cite{KUMAR2017410} discussed an approach for checking
temporal compliance. They used a mathematical
%An approach for checking temporal compliance is discussed in \cite{KUMAR2017410}.This approach uses a mathematical
optimization  model to check for violations.
After a violation occurs it can also check
whether the remaining process instance can be completed without further violations
and determine the best way to do so.
In this way, the level of compliance along the time dimensions can be managed.

%!TEX root = ../manuscript.tex

\section{Conclusions and Future Work}
\label{section:conclusions}

Compliance to policies, rules and regulations  is usually treated in a rigid manner in business, government, and other kinds of organizations.  Compliance pertains to matters that affect employees, customers, and just ordinary citizens. Rigid compliance means that either there is strict adherence to a rule or policy by an entity in which case the entity is compliant, else the entity is treated as being non-compliant or in violation of the rule or policy.  In the real-world, however, such a binary approach is not very efficient because violations related to processing of applications, permits, invoices, fines, taxes, etc. may occur along a continuous spectrum and even minor violations may lead to cancellation of transactions or processes.  Hence, it is important to recognize the extent of the violation and also allow for remediation or compensation mechanisms for them that are commensurate with the degree of the violation.  This would enhance overall social value by reducing inefficiencies and cutting down wasteful work performed in the system.  

In this paper,
we propose notions of full-, partial-
and non-compliance to describe the compliance levels of a business process
during its execution,
and, based on the information available on different  compliance dimensions for the attributes of a task,
we have proposed a metrics-based framework
that can be used to measure the level of compliance 
and provide more information on the state of a \gls{bp} instance during execution
and auditing phases.  The framework was developed from basic principles of partial compliance. 
% and can provide more information on the state-of-affairs of the \gls{bp}beyond the current notions of process compliance.
% and have presented a conceptual framework
% that can be used to quantify the level of compliance of a \gls{bp} during execution
% and auditing phases.

To realize the effectiveness of the proposed framework,
from an implementation perspective,
we are planning to implement it as a ProM Plugin\footnote{ProM Tools: \url{http://www.promtools.org/doku.php}} such that, given a process log,
the application can automatically perform a compliance evaluation and analysis on the log, and generate a full report that shows compliance at multiple levels of aggregation.  We would also like to test, validate and fine tune this tool by applying it to real-world logs, and seeking feedback from the domain experts about the perceived value they obtain from such an analysis. 
%evaluate the level of compliance of a \gls{bp}
%and can predict the performance of the \gls{bp} with the information available using some data analysis techniques.
Further, it would be useful to extend the notions of compensation more formally. 
%Finally, to cover the full spectrum of the business process compliance  challenge,
%we  plan to extend the presented framework to cover the case of \gls{partial compliance} in the design phase.

% The remaining paper is structured as follows:
% next we tersely discuss various classes of norms in \Cref{section:revisitingNormsClasses},
% {\color{red} contract?}
% followed by our proposal on the notion of partial compliance based upon the degree of violation
% and compensation in \Cref{section:partialComplianceFramework}.
% Related works are discussed in \Cref{section:relatedWork} before concluding the paper with some remarks
% and providing pointers future work in \Cref{section:conclusions}.

% \input{contents/relatedWork}
% \input{contents/background}
% \input{contents/partialComplianceFramework}
% \input{contents/discussions}
% \paragraphbreak
% \input{contents/norms}
% \input{contents/businessContract}
% \input{contents/partialComplianceFramework_orig}
% \input{contents/conclusions}

% \input{contents/example_paymentMaking}

\bibliographystyle{Latex/Classes/splncs04}
\bibliography{samplepaper}

% \clearpage

% \begin{figure}\centering
% \resizebox{\linewidth}{!}{\input{images/designTime/complaintHandling}}
% \caption{Complaint handling process}
% \label{figure:compliantHandlingProcess}
% \end{figure}

% \begin{figure}\centering
% \subfloat[Represented using \gls{bpmn} \adoptedfrom{\cite{Hashmi.2015}}\label{figure:paymentProcess:bpmn}]{\resizebox{\linewidth}{!}{\input{images/frag2}}} \\
% \subfloat[Equivalent \gls{wf-net} (with Syntactical Sugar)\label{figure:paymentProcess:wfnet}]{\resizebox{.75\linewidth}{!}{\input{images/designTime/frag2_wfnet}}}
% \caption{Payment process model}
% \label{figure:paymentProcessModel}
% \end{figure}

% \begin{figure*}
% \resizebox{\linewidth}{!}{\input{images/designTime/testProcess}}
% \caption{Test Process}
% \label{figure:testProcess}
% \end{figure*}

% \Cref{figure:testProcess}

% \begin{itemize}
% \item the proposed approach can be potentially used in various domains including Cyber-Phyical and Safety Critical systems.
% \item Combining multiple degree of compliance of deg 1, deg2,
% 	and deg3 \\
% 	$deg1 \& deg2 = deg1 \times deg2 $ \\
% $deg1 \vee deg2 = 1-1-deg1\times 1-deg2$
% \end{itemize}

\end{document}